\documentclass[letterpaper]{article} % DO NOT CHANGE THIS

\usepackage{aaai2026}  % DO NOT CHANGE THIS

\usepackage{times}  % DO NOT CHANGE THIS
\usepackage{helvet}  % DO NOT CHANGE THIS
\usepackage{courier}  % DO NOT CHANGE THIS
\usepackage[hyphens]{url}  % DO NOT CHANGE THIS
\usepackage{graphicx} % DO NOT CHANGE THIS
\usepackage{natbib}  % DO NOT CHANGE THIS AND DO NOT ADD ANY OPTIONS TO IT
\usepackage{caption} % DO NOT CHANGE THIS AND DO NOT ADD ANY OPTIONS TO IT
\usepackage{algorithm}
\usepackage{algorithmic}
\usepackage{xcolor} % 用于颜色支持

\usepackage{newfloat}
\usepackage{listings}
\DeclareCaptionStyle{ruled}{labelfont=normalfont,labelsep=colon,strut=off} % DO NOT CHANGE THIS
\floatstyle{ruled}
\newfloat{listing}{tb}{lst}{}
\floatname{listing}{Listing}
\title{AdaCuRL: Adaptive Curriculum Reinforcement Learning with Invalid \\ Sample Mitigation and Historical Revisiting}
\author{
    Renda Li,
    Hailang Huang,
    Fei Wei\thanks{Project leads and corresponding authors.},
    Feng Xiong,
    Yong Wang\footnotemark[1],
    Xiangxiang Chu
}
\affiliations{
    AMAP, Alibaba Group\\
}
\usepackage{bibentry}
\usepackage{booktabs}
\usepackage{amsmath}
\usepackage{amssymb} 
\usepackage{enumitem}
\usepackage{subcaption}
\usepackage{multirow}
\usepackage{xspace}
\usepackage{adjustbox}
\usepackage{makecell}
\newcommand{\ours}{AdaCuRL\xspace}
\usepackage{tabularx}

\begin{document}

\maketitle
\begin{abstract}
Reinforcement learning (RL) has demonstrated considerable potential for enhancing reasoning in large language models (LLMs). 
However, existing methods suffer from Gradient Starvation and Policy Degradation when training directly on samples with mixed difficulty. To mitigate this, prior approaches leverage Chain-of-Thought (CoT) data, but the construction of high-quality CoT annotations remains labor-intensive. Alternatively, curriculum learning strategies have been explored but frequently encounter challenges, such as difficulty mismatch, reliance on manual curriculum design, and catastrophic forgetting.
To address these issues, we propose \ours, a \textbf{Ada}ptive \textbf{Cu}rriculum \textbf{R}einforcement \textbf{L}earning framework that integrates coarse-to-fine difficulty estimation with adaptive curriculum scheduling. This approach dynamically aligns data difficulty with model capability and incorporates a data revisitation mechanism to mitigate catastrophic forgetting. Furthermore, \ours employs adaptive reference and sparse KL strategies to prevent Policy Degradation. Extensive experiments across diverse reasoning benchmarks demonstrate that \ours consistently achieves significant performance improvements on both LLMs and MLLMs.
\end{abstract}

\section{Introduction}
Post-training methods designed to enhance complex reasoning capabilities have emerged as a prominent area of research. Supervised fine-tuning (SFT) typically distills expert models to obtain high-quality reasoning trajectories for achieving satisfactory performance. In contrast, RL-based approaches, exemplified by GRPO~\cite{guo2025deepseek}, demonstrate that models can self-improve reasoning through RL without relying on high-quality distillation data. This has inspired numerous subsequent efforts on both LLMs \cite{dang2025reinforcementlearningreasoningsmall, chu2025gpg} and MLLMs \cite{chen2025r1v, meng2025mm}.

\begin{figure}[ht]
    \centering
    \includegraphics[width=1.0\linewidth]{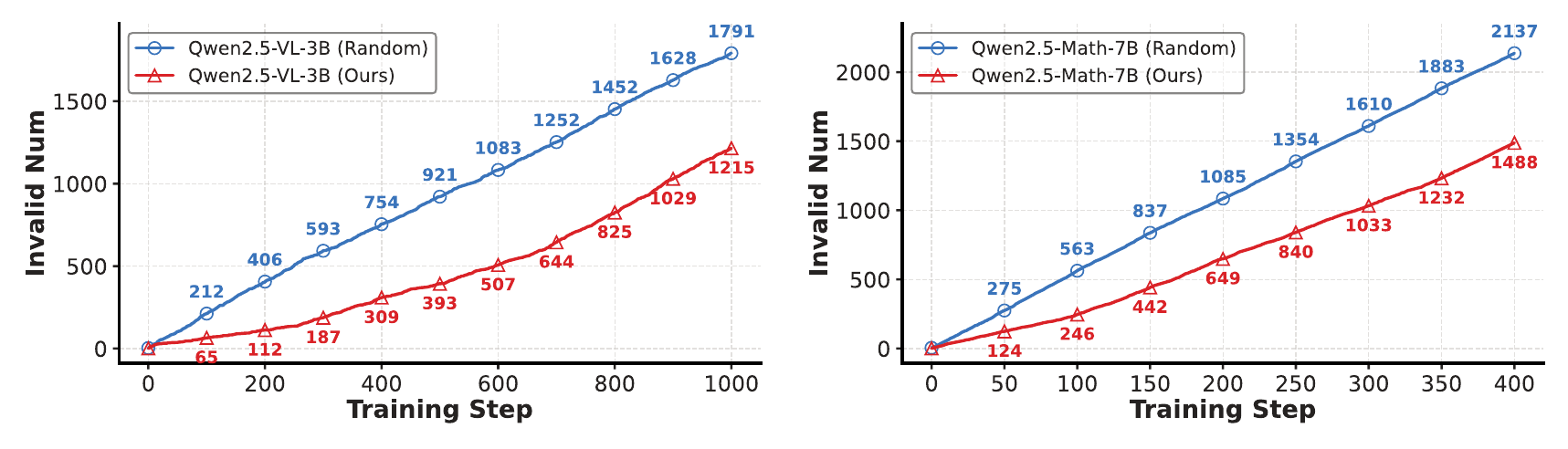}
    \caption{Cumulative invalid samples during GRPO training: shuffled data (Baseline) vs curriculum learning (Ours) on standard open-source datasets.}
    \vspace{-4mm}
    \label{fig:intro}
\end{figure}
Despite the notable success of RL-based methods, their performance is critically dependent on the training data curriculum.
A critical bottleneck emerges when models are trained on mixed-difficulty data, leading to severe \textbf{Data Inefficiency}. 
This inefficiency stems from an intrinsic coupling between sample difficulty and the relative advantages of rollouts within GRPO groups.
Specifically, when training samples exhibit extreme difficulty levels relative to the current policy, the reward signal often collapses into a binary state, where simple samples uniformly receive rewards of 1, while difficult ones invariably yield rewards of 0 \cite{chu2025gpg}. 
These invalid samples culminate in two core dilemmas: \textbf{Gradient Starvation} and \textbf{Policy Degradation}.
(i) Gradient Starvation occurs when all exploration rollouts produce rewards of 1 or 0, causing the advantage function to collapse to zero. Consequently, the policy gradient is nullified, depriving the model of any meaningful learning signal. (ii) Policy Degradation arises when the KL divergence penalty imposed on invalid samples dominates the optimization signals. This forces the policy to revert to a conservative reference model, which impairs the reasoning capability obtained through RL.
As illustrated in Figure~\ref{fig:intro}, these invalid samples are common in standard datasets, highlighting the urgent need for an improved training paradigm.

To alleviate Gradient Starvation, it is necessary to avoid rollouts producing all-zero or all-one rewards.
Hint-GRPO~\cite{huang2025boosting} incorporates expert reasoning trajectories to avoid difficult samples receiving all-zero rewards. 
However, this approach does not recognize the importance of aligning model capability and sample difficulty during the RL process.
The methods based on curriculum learning aim to train on samples matched to the model's capacity to reduce invalid samples. However, existing approaches face three primary limitations: (i) difficulty mismatch, (ii) manual curricula, and (iii) forgetting of past samples. Some works define difficulty using human prior knowledge \cite{deng2025boosting, song2025fastcurl} or expert models \cite{shi2025efficient}, while failing to capture how the model itself perceives difficulty.
Additionally, other works \cite{team2025kimi, deng2025boosting} employ handcrafted curricula as training schedulers, without considering model feedback. Moreover, all the above methods typically train within a specific difficulty range, neglecting earlier data, and as training shifts toward harder samples, performance on easier data may deteriorate.
Although the above methods reduce the frequency of invalid samples, they lack effective mechanisms for addressing Policy Degradation caused by the occurrence of invalid samples.

To address the above issues, we propose \ours, a novel curriculum reinforcement learning approach. 
Specifically, \ours introduces a coarse-to-fine difficulty estimation strategy that can sample data with desired difficulty distributions from large-scale datasets and accurately estimate sample difficulty.
During fine-tuning, \ours partitions the data into buckets from easy to hard and updates the specific bucket based on model feedback to avoid invalid samples. 
Besides, the bucket update mechanism allows revisiting historical data to mitigate catastrophic forgetting.
To prevent degradation from invalid samples, \ours incorporates a sparse KL mechanism.
Furthermore, we introduce self-pacing mechanism into \ours, called Re-\ours, which enhances data utilization and continuously improves reasoning.

Our main contributions are as follows:
\begin{itemize}
\item We propose \ours, which integrates a coarse-to-fine difficulty estimation strategy and a novel curriculum RL algorithm to enhance data efficiency in GRPO, enabling (M)LLMs to progressively improve their reasoning capabilities.
\item We design sparse KL to effectively prevent Policy Degradation and propose an adaptive reference strategy to avoid excessive alignment with the reference model.
\item We further introduce Re-\ours, which iteratively re-estimates sample difficulty and conducts curriculum RL to mine data and strengthen reasoning.
Extensive experiments across diverse benchmarks demonstrate the effectiveness of our approach on both MLLMs and LLMs.
\end{itemize}

\section{Preliminary}
\textbf{Curriculum Learning (CL)} is a training strategy inspired by human incremental learning, where models are trained sequentially on data ordered by difficulty to achieve better performance and faster convergence. 
CL comprises two main components, difficulty estimation and a training scheduler, and both can be categorized as either predefined or automatic.
Formally, given a dataset:
\begin{equation}
    \mathcal{D} = \{(x_i, y_i, d_i)\}_{i=1}^{N},
\end{equation}
where \( d_i \) is the difficulty of sample \((x_i, y_i)\) and \(N\) is the dataset size, the curriculum ensures: \(d_1 \leq d_2 \leq \dots \leq d_N\).

% \cxx{What's the meaning of N?  }

Self-Paced Learning (SPL) automates difficulty estimation by selecting samples according to current loss. Formally, SPL constructs training sets for each epoch as:
\begin{equation}
    \mathcal{D} = \{(x_i,y_i)\mid \ell(f_{\mathbf{w}}(x_i), y_i)\leq\tau\},
\end{equation}
where \(\tau\) is an adaptive loss threshold.

Predefined schedulers select samples according to fixed rules \cite{dong2025insight}, whereas automatic ones~\cite{graves2017automated} make scheduling decisions based on the model’s feedback.

\noindent\textbf{Group Relative Policy Optimization (GRPO)} eliminates the need for a value model by normalising outcome rewards within a group of \(G\) samples and applying a policy-gradient objective regularised by a KL term.

For a prompt \(q\), the policy \(\pi_\theta\) generates \(G\) responses \(\{o_i\}\) with scalar rewards \(\{r_i\}\). Let \(\mu_r\) and \(\sigma_r\) denote the group mean and standard deviation. GRPO defines the group-relative advantage: \(\hat{A}_i = \frac{r_i-\mu_r}{\sigma_r+\varepsilon}\), where \(\varepsilon>0\) prevents division by zero.
We define \(\rho_i = \pi_\theta(o_i\mid q)\big/\pi_{\text{old}}(o_i\mid q)\) as the importance ratio between the learned policy $\pi_\theta$ and a fixed reference policy $\pi_{ref}$, and \(\text{clip}(\rho_i, 1-\epsilon, 1+\epsilon)\) as the CLIP operation.
The objective of GRPO is then expressed as:
\begin{equation}
\begin{split}
    \mathcal{L}_{\text{GRPO}}(\theta)
    &= -\mathbb{E}_{i}\!\bigl[\min(\rho_i \cdot \hat{A}_i, \text{CLIP} \cdot \hat{A}_i)\bigr] \\
    &\quad + \beta\,\mathbb{E}_{i}\!\Bigl[\mathrm{KL}\!\bigl(\pi_\theta \,\|\, \pi_{\text{ref}}\bigr)\Bigr],
\end{split}
\label{eq:grpo-loss}
\end{equation}
where \(\beta\) controls the KL regularization strength.

\begin{figure*}[ht]
  \centering
  \includegraphics[width=\linewidth]{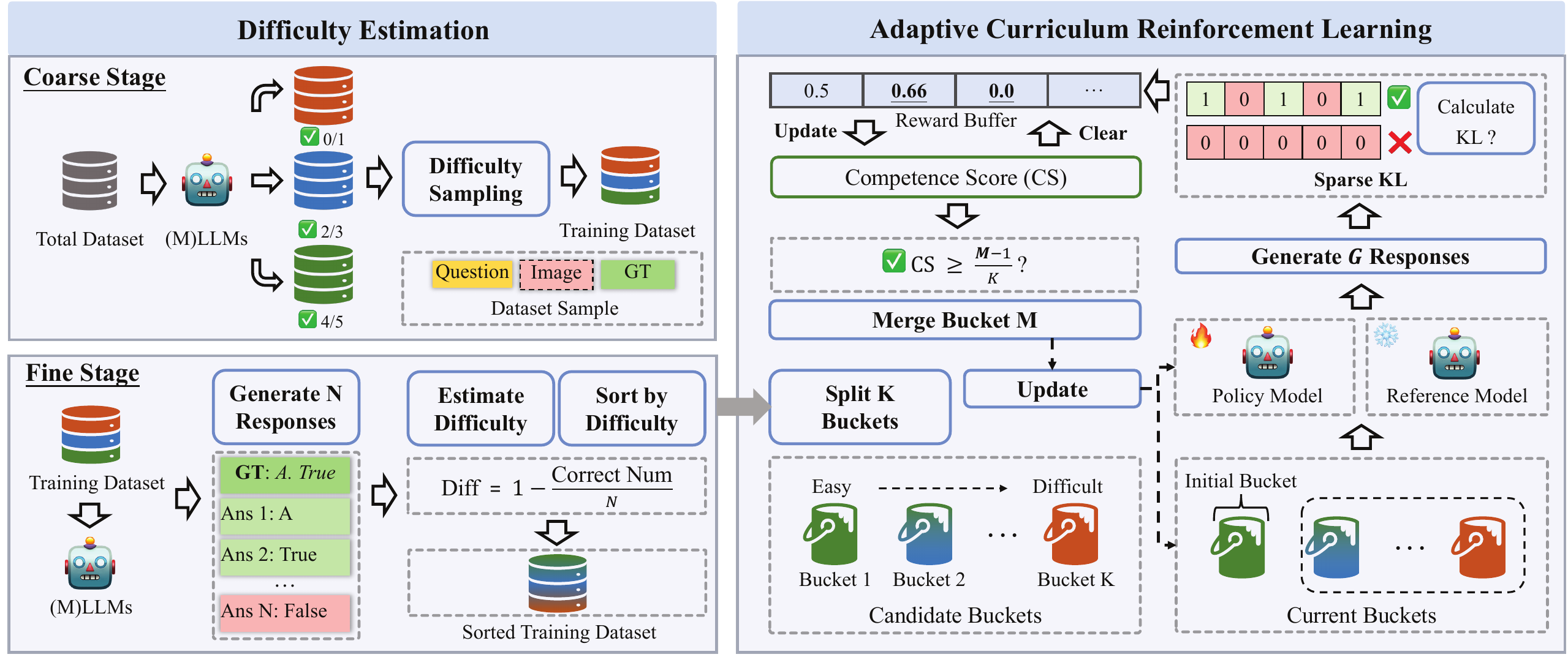}
  \caption{
  The overall framework of \ours.
Difficulty Estimation (left) samples a training subset from a large-scale dataset to match a target difficulty distribution and sorts the data from easy to hard.
Curriculum Reinforcement Learning (right) monitors the average accuracy reward during training to assess the model’s mastery of the current difficulty level and progressively introduces more challenging samples.
In addition, \ours incorporates sparse KL and adaptive reference mechanisms to prevent degradation of the model’s reasoning capability.}
  \label{fig:method}
\end{figure*}

\section{Method}
\ours consists of three key components. First, we introduce a coarse-to-fine difficulty estimation strategy to effectively extract subsets with a target difficulty distribution from large-scale datasets. Then, we present the core training scheduling algorithm, which serves as the central framework of \ours. Finally, we extend this framework with Re-\ours, an enhanced variant designed to further optimize data utilization for improved reasoning capabilities. The detailed algorithm is provided in the Appendix \ref{app_algo}.

\subsection{Coarse-to-Fine Difficulty Estimation}
\label{sec:coarse_to_fine}

Accurate difficulty estimation is essential for effective curriculum learning. 
We adopt an unbiased approach to evaluate problem difficulty based on the frequency of correct solutions generated by the base model across multiple attempts~\cite{snell2024scaling, shi2025efficient}.

Curriculum learning typically requires sampling from the training set to form a specified difficulty distribution (e.g., containing more hard problems than easy ones). However, given the large dataset size, precisely estimating each sample’s difficulty by generating answers multiple times incurs substantial inference overhead, while random sampling fails to match the desired difficulty distribution. To address this, we propose a coarse-to-fine difficulty estimation strategy.

\begin{itemize}
    \item \textbf{Coarse stage.} For each problem, the model produces five answers. Based on the number of correct answers, we assign each problem to one of three bins. We then sample from each bin according to a predefined ratio, while ensuring that the selected samples remain evenly distributed across datasets. Formally, let $c_i \in \{0, \dots, 5\}$ be the number of correct answers for problem $i$. We define three bins: $\mathcal{G}_1 = \{i \mid c_i \in \{0, 1\}\}$, $\mathcal{G}_2 = \{i \mid c_i \in \{2, 3\}\}$, and $\mathcal{G}_3 = \{i \mid c_i \in \{4, 5\}\}$,
    and draw
    \begin{equation}
        \label{eq:sampling}
        \mathcal{S}\;=\;
        \bigcup_{k=1}^{3}
        \operatorname{Sample}\!\Bigl(
          \mathcal{G}_k,\;
          n_k=\lfloor \rho_k\,\lvert \mathcal{G}_k\rvert \rfloor
        \Bigr),
    \end{equation}
    where \(\rho_k\) is the predefined sampling ratio for \(k\)-th bin.

    \item \textbf{Fine stage.} For each problem in $\mathcal{S}$, we generate \(N\) (\(N \gg 5\)) answers for precise difficulty estimation. Let \(c(q)\) denote the number of correct solutions out of these \(N\) attempts for problem~\(q\). We define its difficulty score as
    \begin{equation}
        \label{eq:difficulty}
        \text{Difficulty}(q) \;=\; 1 - \frac{c(q)}{N},
    \end{equation}
    and then filter out problems with difficulty above 0.95 or below 0.05 to avoid overly hard or trivial cases. The remaining data are sorted by ascending difficulty to form the final training dataset~\(\mathcal{D}\).
\end{itemize}

\subsection{Curriculum Reinforcement Learning}
\label{sec:cl}
After sorting \(\mathcal{D}\) by difficulty, we partition it into \(K\) consecutive buckets \(\{\mathcal{B}1,\mathcal{B}_2,\dots,\mathcal{B}_K\}\) with equal size:
\begin{equation}
  \label{eq:bucket-def}
  \mathcal{B}_k =
  \Bigl\{
    q_{(k-1)\frac{|\mathcal{D}|}{K}+1},
    \dots,
    q_{\,k\frac{|\mathcal{D}|}{K}}
  \Bigr\},\quad
  k = 1,\dots,K,
\end{equation}
where \(q_1,\dots,q_{|\mathcal{D}|}\) are ordered from easy to hard.

The current training subset \(\mathcal{D}_{\mathrm c}\) is initialized as the first bucket \(\mathcal{B}_1\). During training, we merge the next bucket into \(\mathcal{D}_{\mathrm c}\) at each update stage \(t\) to mitigate catastrophic forgetting:
\begin{equation}
  \label{eq:baby-step}
  \mathcal{D}_{\mathrm c}^{(t+1)} =
  \mathcal{D}_{\mathrm c}^{(t)} \cup \mathcal{B}_{t+2},
  \quad t = 0,\dots,K-2.
\end{equation}
The training processs continues until \(\mathcal{D}_{\mathrm c} = \mathcal{D}\).
This incremental expansion retains knowledge from easier samples during initial stages, while gradually adding more challenging samples as the model's reasoning capabilities improve.

\subsubsection{Reward Function.}
We use two binary reward signals: \emph{format reward} and \emph{accuracy reward}. We observe that the format reward converges rapidly, while the accuracy reward, especially from harder buckets, remains relatively low and progresses slowly. This imbalance affects the advantage function in Equation~\eqref{eq:grpo-loss}, dominated by the format reward and hindering \(\pi_\theta\) from learning accurate reasoning paths effectively. To address this, we update the policy solely based on the accuracy reward after \(T_f\) training steps.

\subsubsection{Bucket Update Strategy.}
A \emph{naive} bucket update strategy trains each bucket sequentially.
Such a schedule is often inefficient and sub-optimal because it ignores the \emph{current state} of the model, leading to over-training of easy buckets while hard buckets may receive insufficient updates.

To adaptively assess progress, we use the \emph{accuracy reward} during training to measure how well the model has mastered the current bucket and record the reward of each sample in the rewards buffer $\mathcal{R}_b$.
% \cxx{What is $\mathcal{R}s$?}  
Specifically, we maintain a \emph{competence score} \(cs\in[0,1]\), which is initialized as \(cs^{(0)} = 0\) and updated as:
\begin{equation}
  \label{eq:grasp-update}
  cs^{(t+1)} \;\leftarrow\;
  cs^{(t)} + \left(\bar{r} - 0.5\right) \times
           \max\left(1 - cs^{(t)},\; \gamma\right)
\end{equation}
where \(\bar{r}\) is the average reward over the most recent
\(M\) training samples and
\(\max(1-cs,\gamma)\) acts as a decay factor on the update rate.
As \(cs\) increases, the update step becomes smaller, mimicking human learning by spending more time on harder buckets, while \(\gamma\)
prevents the rate from becoming too small.

Once $\mathcal{R}_b$ contains \(M\) samples, we update \(cs\) and check whether the curriculum set \(\mathcal{D}_{\mathrm c}\) should be expanded. 
% \cxx{What is $M$?} 
The curriculum expansion condition is defined as follows:
\begin{equation}
    cs \;\ge\; \frac{k - 1}{K}
    \label{eq:grasp-threshold}
\end{equation}
When the condition in Equation~\eqref{eq:grasp-threshold} is satisfied for the next bucket index \(k\), bucket \(\mathcal{B}_k\) is merged into \(\mathcal{D}_{\mathrm c}\).

% \cxx{If the problem is too difficult, how can you guarantee cs is increased to meet Eq 11?}
To keep the estimate of \(\bar{r}\) faithful to the model’s ability on
\emph{newly introduced} data, only samples drawn from the \emph{latest
merged bucket} contribute to \(\bar{r}\).
Upon merging $\mathcal{B}_k$, the competence score is re-initialized to $cs = \tfrac{k - 1}{K}$ to ensure an accurate reflection of the policy model’s mastery over the data in the newly added bucket.

\subsubsection{KL Divergence Design.}
In Equation~\eqref{eq:grpo-loss}, GRPO calculates the KL divergence with the base model during each loss computation, leading to two issues: (i) When the advantage function is a full zero vector, the loss is dominated by the KL term, causing the policy model to unnecessarily align with the base model, (ii) as the model's reasoning ability improves, continuing to compute KL divergence with the base model undermines the already acquired reasoning capabilities. To address these limitations, we introduce two strategies into our proposed framework:

\begin{itemize}[leftmargin=*]
    \item \textbf{Conditional KL computation}. When all rewards within a rollout group are either 0 or 1, we exclude the KL divergence term from the loss computation for that specific group, enabling more effective enhancement of the model's reasoning abilities. The GRPO loss in \ours is defined as follows:
    \begin{equation}
        \begin{aligned}
        \mathcal{L}_{\text{GRPO}}(\theta)
        &= -\,\mathbb{E}_{i}\!\bigl[\min(\rho_i \cdot \hat{A}_i, \text{CLIP} \cdot \hat{A}_i)\bigr]\\
        &\quad +\,
          \mathbb{I}\!\bigl[\hat{A}_i \neq 0\bigr]\,
          \beta\,
          \mathbb{E}_{i}\!\Bigl[
            \mathrm{KL}\!\bigl(\pi_\theta \,\|\, \pi_{\text{ref}}\bigr)
          \Bigr]
        \end{aligned}
        \label{eq:grpo-loss_ours_conditional}
    \end{equation}

    \item \textbf{Reference model resetting}.
    After each bucket update, the reference model \(\pi_\text{ref}\) is reset to the current policy model \(\pi_\theta\), thus avoiding excessive alignment with the initial reference model as the reasoning capability of \(\pi_\theta\) improves.

\end{itemize}

\subsection{Self-pacing Mechanism}
\label{sec:selfpacing}

After the first round of training with coarse-to-fine difficulty estimation and curriculum RL, the model develops stronger reasoning capabilities. To further improve performance, we introduce a \emph{self-pacing} mechanism, called Re-\ours.

Specifically, we refine the coarse-to-fine difficulty estimation using the updated policy model \(\pi_\theta\) and filter out previously trained data during sampling.
Let \(\text{Difficulty}^{(1)}(q)\) denote the re-estimated difficulty score.
To preserve acquired reasoning capabilities, we discard samples with difficulty scores below a threshold (e.g., 0.2) in the second iteration:
\begin{equation}
  \label{eq:filter-easy}
  \mathcal{D}' = \{\, q \in \mathcal{D} \mid \text{Difficulty}^{(1)}(q) \ge 0.2 \,\} .
\end{equation}

The remaining data \(\mathcal{D}'\) is then sorted and repartitioned into \(K\) buckets in ascending order of updated difficulty:
\begin{equation}
  \label{eq:bucket-selfpacing}
  \mathcal{B}_k^{(1)} =
  \Bigl\{
    q_{(k-1)\frac{|\mathcal{D}'|}{K}+1}^{(1)},
    \dots,
    q_{k\frac{|\mathcal{D}'|}{K}}^{(1)}
  \Bigr\},\quad
  k = 1,\dots,K .
\end{equation}

We then repeat the training process described in Sec.~\ref{sec:cl} on these updated buckets. 
This self-pacing mechanism allows data that was previously excluded due to excessive difficulty to be revisited, while simultaneously filtering out samples already solved with high confidence. 
As a result, the current policy \(\pi_\theta\) continues to train on increasingly informative data, further enhancing its reasoning capabilities.

\section{Experiments}

\subsection{Datasets}
\label{sec:dataset}
For training MLLMs, we curate a training dataset from a broad range of mathematical reasoning sources, including CLEVR~\cite{johnson2017clevr}, CLEVR\text{-}Math~\cite{lindstrom2022clevr}, Geo3K~\cite{lu2021inter}, GeoMverse~\cite{kazemi2023geomverse}, GeoQA+~\cite{chen2021geoqa}, IconQA~\cite{lu2021iconqa}, Super\text{-}CLEVR~\cite{li2023super}, TabMWP~\cite{lu2022dynamic}, UniGeo~\cite{chen2022unigeo}, GEOS\cite{seo2015solving}, WeMath~\cite{qiao2024we}, SceMQA~\cite{liang2024scemqa}, and PolyMath~\cite{gupta2024polymath}. Together, they comprise about 100K problems spanning various types (e.g., geometry, algebra, counting) and difficulty levels.
To align with GRPO, we filter out samples whose answers cannot be reliably validated. For multiple-choice questions, we standardize the format to explicitly include both the option label and content (e.g., ``A. 1.8''), preventing the model from exploiting superficial answer patterns.
As detailed in Sec.~\ref{sec:coarse_to_fine}, we partition the data into three coarse difficulty groups ($\mathcal{G}_1$, $\mathcal{G}_2$, and $\mathcal{G}_3$). Following standard curriculum learning, we increase the proportion of harder samples by sampling 2K, 3K, and 5K examples from these groups, yielding a 10K training dataset.

For training LLMs, we utilize the Open-RS dataset~\cite{dang2025reinforcementlearningreasoningsmall}, which contains 7K samples. Given its moderate size, we directly perform fine-grained difficulty estimation and sorting.

\begin{table*}[ht]
    \centering
    \small
    \setlength{\tabcolsep}{0.7mm}    
    \begin{tabular}{l|cccccc|cccccc}
        \toprule
        \multirow{3}{*}{\textbf{Model}} & \multicolumn{6}{c|}{\textbf{Mathematical Reasoning}} & \multicolumn{6}{c}{\textbf{General Reasoning}} \\
        \cmidrule(l){2-13}
        & \textbf{DynaMath} & \textbf{MathVista} & \textbf{Math-V} & \textbf{MathVerse} & \textbf{LogicVista} & \textbf{Avg.} 
        & \textbf{MMStar} & \textbf{MMMU} & \textbf{Hallu.} & \textbf{AI2D} & \textbf{MMVET} & \textbf{Avg.} \\
        \midrule
        \multicolumn{13}{c}{\textbf{\textit{Qwen2.5-VL-3B Models}}} \\
        \midrule
        Qwen2.5-VL-3B & 40.90 & 62.00 & 22.62 & 33.75 & 38.70 & 39.59
                               & 56.00 & 50.88 & 45.66 & 80.40 & 60.20 & 58.63 \\
        + SFT                  & 38.74    & 60.60    & 22.27    & 34.37    & 41.61    & 39.52   
                               & 58.00    & 51.11    & 49.88    & 79.60    & 63.71    & 60.46   \\
        + GRPO                 & 41.16 & 65.00 & 23.02 & 35.31 & 38.70 & 40.64
                               & 55.53 & 52.11 & 47.14 & 77.95 & 61.37 & 58.82 \\

        \midrule
        + \ours (Easy)           & 45.44 & 64.10 & 22.10 & 37.00 & 39.37 & 41.60
                               & 57.60 & 52.00 & \textbf{50.58} & 81.60 & 61.78 & 60.71 \\
        + \ours (Hard)           & 42.43 & 66.20 & 22.56 & 35.96 & 38.92 & 41.21
                               & 58.66 & 52.33 & 46.76 & 78.17 & 60.36 & 59.26 \\
        + \ours                  & 48.10 & 66.50 & 23.70 & 40.67 & 40.09 & 43.81
                               & 59.95 & 52.66 & 49.03 & 81.34 & 62.76 & 61.15 \\
        + Re-\ours    & \textbf{49.22} & \textbf{67.40} & \textbf{24.54} & \textbf{42.24} & \textbf{42.51} & \textbf{45.18}
                               & \textbf{60.07} & \textbf{53.11} & 48.27 & \textbf{81.74} & \textbf{63.64} & \textbf{61.37} \\
        \midrule
        \multicolumn{13}{c}{\textbf{\textit{Qwen2.5-VL-7B Models}}} \\
        \midrule        
        Qwen2.5-VL-7B & 51.99 & 68.50 & 25.42 & 44.53 & 46.97 & 47.48
                               & 65.00 & 58.22 & 52.35 & 84.71 & 67.38 & 65.53 \\
        + SFT                  & 44.59    & 64.20    & 39.69    & 25.59    & 43.62    & 43.54   
                               & 62.93    & 56.00    & 52.72    & 83.45    & 64.86    & 63.99   \\
        + GRPO                 & 48.12    & 70.90    & 26.94    & 47.22    & 45.41    & 47.72   
                               & 63.06    & 57.44    & 54.42    & 83.29    & 69.03    & 65.45   \\

        \midrule
        + \ours                  & 55.10    & 70.40    & 27.07    & \textbf{48.75}    & 48.10    & 49.88   
                               & \textbf{65.36}    & \textbf{58.66}    & \textbf{57.27}    & \textbf{85.85}    & 69.31    & \textbf{67.29}   \\
        + Re-\ours    & \textbf{56.67}    & \textbf{71.60}    & \textbf{28.92}    & 48.38    & \textbf{48.99}    & \textbf{50.91}   
                               & 65.27    & 58.00    & 56.53    & 85.56    & \textbf{69.91}    & 67.05   \\
        
        \bottomrule
        
    \end{tabular}
    \caption{Comparison of methods on mathematical (Left) and general (Right) reasoning benchmarks for MLLMs.}
    \label{tab:main_result}
\end{table*}

\begin{table}[ht]
  \centering
  \small                           
  \setlength{\tabcolsep}{0mm}
  \begin{tabularx}{\linewidth}{l*{6}{>{\centering\arraybackslash}X}}
    \toprule
    \textbf{Model} &
    \textbf{AIME} & \textbf{AMC} & \textbf{MATH} &
    \textbf{Minerva} & \textbf{Oly.} & \textbf{Avg.} \\
    \midrule
    \multicolumn{7}{c}{\textbf{\textit{Qwen2.5-Math-1.5B Models}}} \\
    \midrule
    Base Model      & 6.45  & 36.40 & 46.33 & 12.62 & 24.74 & 25.31 \\
    + GRPO          & 7.50  & 40.62 & 56.00 & 12.99 & 27.25 & 28.87 \\
    \midrule
    + \ours         & \textbf{9.58}  & 45.63 & \textbf{62.46} & \textbf{14.58} & 29.33 & \textbf{32.32} \\
    \quad- SparseKL & 9.29  & \textbf{45.71} & 61.46 & 14.46 & \textbf{29.53} & 32.09 \\
    \quad- Reset Ref& 9.37  & 45.00 & 59.13 & 14.34 & 28.74 & 31.32 \\
    \quad- Revisiting & 8.13 & 44.22 & 60.46 & 13.60 & 29.18 & 31.12 \\
    \midrule
    \multicolumn{7}{c}{\textbf{\textit{Qwen2.5-Math-7B Models}}} \\
    \midrule
    Base Model      & 15.83 & 51.87 & 64.66 & 17.40 & 29.18 & 35.79 \\
    + GRPO          & 18.95 & 56.56 & 68.80 & 17.28 & 31.55 & 38.63 \\
    \midrule
    + \ours         & \textbf{22.22} & \textbf{59.22} & \textbf{74.53} &
                      \textbf{27.33} & \textbf{37.48} & \textbf{44.16} \\
    \bottomrule
  \end{tabularx}
  \caption{
    Comparison of methods on mathematical reasoning benchmarks for LLMs. Results averaged over AIME24@16, AMC23@16, others@3. Oly.\ denotes Olympiad-bench.
    }
  \label{tab:main}
\end{table}

\subsection{Benchmarks}
For MLLMs, we build two complementary benchmarks: mathematical reasoning and general multimodal reasoning. For LLMs, we adopt standard mathematical reasoning benchmarks.

The multimodal mathematical reasoning benchmark comprises DynaMath~\cite{zou2024dynamath}, MathVista\_MINI~\cite{lu2023mathvista}, Math-V~\cite{wang2024measuring}, MathVerse\_MINI~\cite{zhang2024mathverse}, and LogicVista~\cite{xiao2024logicvista}, and the multimodal general reasoning benchmark includes MMStar~\cite{chen2024we}, MMMU~\cite{yue2024mmmu}, HallusionBench~\cite{guan2024hallusionbench}, AI2D~\cite{kembhavi2016diagram}, and MMVET~\cite{yu2023mm}. For unimodal reasoning, we adopt standard datasets such as AIME24, AMC23, MATH500~\cite{lightman2023let}, Minerva~\cite{lewkowycz2022solving}, and Olympiadbench~\cite{he2024olympiadbench}. Together, these benchmarks offer a comprehensive, multi-dimensional assessment of the models’ reasoning capabilities.

\subsection{Training Settings}
We employ Qwen2.5-VL-3B-Instruct~\cite{Qwen2.5-VL} and Qwen2.5-VL-7B-Instruct for multimodal experiments. For fine-grained difficulty estimation, we set $N = 100$ generations, format reward cutoff $T_f = 64$, decay $\gamma = 0.5$, and competence score interval $M = 512$. Unless specified, $K = 4$ buckets are used.
For unimodal experiments, we use Qwen2.5-Math-1.5B~\cite{yang2024qwen2} and Qwen2.5-Math-7B as base models. 
Unlike Open-RS's cosine reward, we employ accuracy reward, maintaining consistency in other hyperparameters. 
We set $K = 3$ buckets, with other curriculum learning hyperparameters following the multimodal settings. 
We train \ours and baselines with same steps and evaluate on the final checkpoint. More details are provided in the Appendix \ref{app_hyp}.

\subsection{Main Results}
Tables~\ref{tab:main_result} and~\ref{tab:main} present a comprehensive comparison of different methods across reasoning benchmarks on both multimodal and language models. The results are as follows.

\noindent\textbf{Neither the original GRPO nor SFT significantly enhances reasoning capabilities.} As shown in Table~\ref{tab:main_result}, the original GRPO improves mathematical and general reasoning by only 0.85\% and 0.19\%, respectively, on Qwen2.5-VL-3B, with similar results on the 7B model. The SFT baseline even leads to degraded performance, particularly on the larger 7B model. We hypothesize this degradation stems from fine-tuning on lower-quality open-source data, which may harm an already strong baseline. For language models, the original GRPO yields noticeable gains, improving by 3.56\% on Qwen2.5-Math-1.5B and 2.84\% on Qwen2.5-Math-7B (Table~\ref{tab:main}). We attribute this to the additional information fusion in multimodal models, which increases the difficulty of reinforcement fine-tuning.

\noindent\textbf{\ours achieves outstanding performance.} On both multimodal and language models, \ours outperforms baselines across all benchmarks and model sizes. For example, on mathematical reasoning, \ours improves by 3.17\% and 2.16\% on Qwen2.5-VL-3B and 7B, respectively, and achieves gains of 3.45\% and 5.53\% on Qwen2.5-Math-1.5B and 7B. These results highlight the importance of progressively increasing training difficulty to enhance reasoning and demonstrate the consistent applicability of \ours to both unimodal and multimodal tasks.

\begin{table}[ht]
  \centering
  \small
  \setlength{\tabcolsep}{1mm}
  \begin{tabularx}{\linewidth}{
    >{\raggedright\arraybackslash}X  
    >{\raggedright\arraybackslash}X  
    *{5}{>{\centering\arraybackslash}X} 
  }
    \toprule
    \textbf{Group} & \textbf{Stage} &
    \begin{tabular}{@{}c@{}}\textbf{clever}\\\textbf{math}\end{tabular} &
    \begin{tabular}{@{}c@{}}\textbf{geo}\\\textbf{3k}\end{tabular} &
    \begin{tabular}{@{}c@{}}\textbf{geom}\\\textbf{verse}\end{tabular} &
    \begin{tabular}{@{}c@{}}\textbf{geoqa}\\\textbf{plus}\end{tabular} &
    \begin{tabular}{@{}c@{}}\textbf{icon}\\\textbf{qa}\end{tabular} \\
    \midrule
    \multirow{2}{*}{$\mathcal{G}_1$}
      & Before & 1142 & 1324 & 558 & 25678 & 3318 \\
      & After  & 849  & 1017 & 451 & 19221 & 2203 \\
    \midrule
    \multirow{2}{*}{$\mathcal{G}_2$}
      & Before & 246  & 926  & 554 & 19502 & 6824 \\
      & After  & 135  & 985  & 408 & 19871 & 4050 \\
    \midrule
    \multirow{2}{*}{$\mathcal{G}_3$}
      & Before & 1560 & 151  & 574 & 3880  & 10423 \\
      & After  & 1964 & 399  & 827 & 9968  & 14312 \\
    \bottomrule
  \end{tabularx}
  \caption{Coarse-grained data distribution before and after one round of training with \ours.}
  \label{tab:data_shift}
\end{table}

\noindent\textbf{Re-\ours further improves reasoning.} As shown in Table~\ref{tab:main_result}, Re-\ours achieves additional improvements of 1.37\% and 1.03\% on mathematical reasoning for the Qwen2.5-VL-3B and 7B models, respectively. 
We further elaborate on the motivation for this approach. Table~\ref{tab:data_shift} shows a shift toward the easier end: samples in $\mathcal{G}_1$ decrease while those in $\mathcal{G}_3$ increase, indicating improved mathematical reasoning capabilities. To further leverage the dataset, we resample after re-estimating difficulty using the updated policy and continue training on the resampled data. 

\begin{figure}[ht]
    \centering
    \includegraphics[width=1.0\linewidth]
    {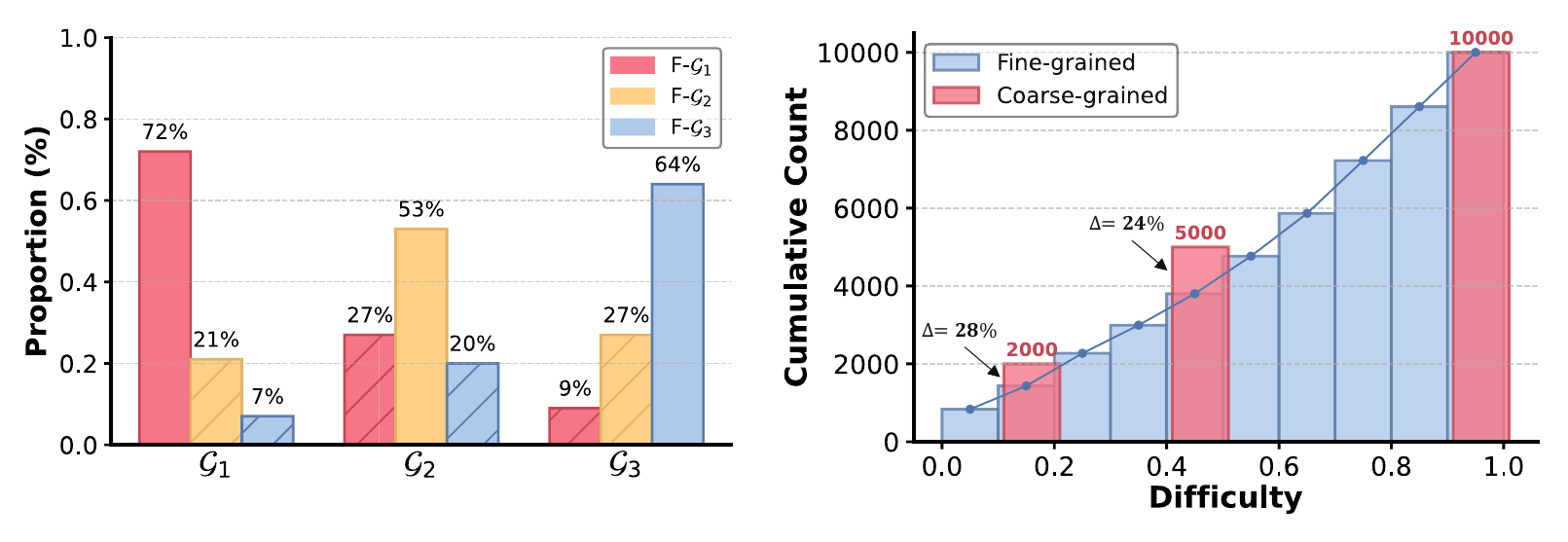}
    \caption{(Left) The proportion of samples from each of the three coarse-grained groups ($\mathcal{G}1/\mathcal{G}_2/\mathcal{G}_3$) that fall into each of the three fine-grained groups (F-$\mathcal{G}_1/\mathcal{G}_2/\mathcal{G}_3$) after fine-grained estimation. 
(Right) The difficulty distribution of coarse-grained sampling compared to that after fine-grained difficulty estimation.}
    \label{fig:data_ana}
\end{figure}

\section{Analysis}
\subsection{Evaluation of Difficulty Estimation}
In this section, we evaluate whether the final difficulty distribution from fine-grained difficulty estimation aligns with the desired coarse-grained sampling distribution. As shown in the right part of Figure~\ref{fig:data_ana}, the red regions represent the desired coarse-grained sampling distribution (i.e., $2\mathrm{K}/3\mathrm{K}/5\mathrm{K}$), while the blue regions indicate the actual distribution obtained through fine-grained estimation. It is evident that the distributions are generally consistent.  
This demonstrates that the proposed coarse-to-fine difficulty estimation method achieves the desired difficulty distribution without large-scale inference.
To further analyze, we use the fine-grained results as ground truth and evaluate the accuracy of the coarse-grained estimation by examining the proportion of fine-grained results within each coarse-grained group, as shown in the left part of Figure~\ref{fig:data_ana}. The accuracy rates for the three groups $\mathcal{G}_1$, $\mathcal{G}_2$, and $\mathcal{G}_3$ are $72\%$, $53\%$, and $64\%$, respectively. This indicates that coarse-grained estimation cannot provide precise difficulty assessments but effectively serves to obtain the desired difficulty distribution.

\begin{table}[ht]
    \centering
    \small
    \setlength{\tabcolsep}{1mm}    
    \begin{tabular}{l c c c c c c }
        \toprule
        \multirow{2}{*}{\textbf{Model}} & \textbf{Dyna} & \textbf{Math} & \textbf{Math-} & \textbf{Math} & \textbf{Logic} & \multirow{2}{*}{\textbf{Avg.}} \\
        & \textbf{Math} & \textbf{Vista} & \textbf{V} & \textbf{Verse} & \textbf{Vista} & \\
        \midrule
        3B + Fine & \textbf{48.10} & \textbf{66.50} & 23.70 & \textbf{40.67} & 40.09 & \textbf{43.81} \\
        \midrule
        3B + Coarse & 45.81  & 65.6  & 22.43 & 38.40  & 41.83 &  42.81 \\
        7B + Fine & 47.84  & 63.9  & \textbf{24.83}  & 38.82  & \textbf{42.73}  & 43.62 \\

        \bottomrule 
    \end{tabular}
        \caption{Results of training with different difficulty estimation strategies using Qwen2.5-VL-3B.} 
    \label{tab:diff_estimation}
\end{table}

\subsection{Different Difficulty Estimation Strategies}
 In addition to the fine-grained difficulty estimation based on the model itself, we explored two alternatives: (i) coarse-grained estimation only, and (ii) fine-grained estimation from a stronger external model. As shown in Table~\ref{tab:diff_estimation}, both yield suboptimal results, highlighting that \ours relies on the model’s own fine-grained estimation to provide accurate difficulty assessments for curriculum learning.

\begin{figure}[ht]
    \centering
    \includegraphics[width=1.0\linewidth]{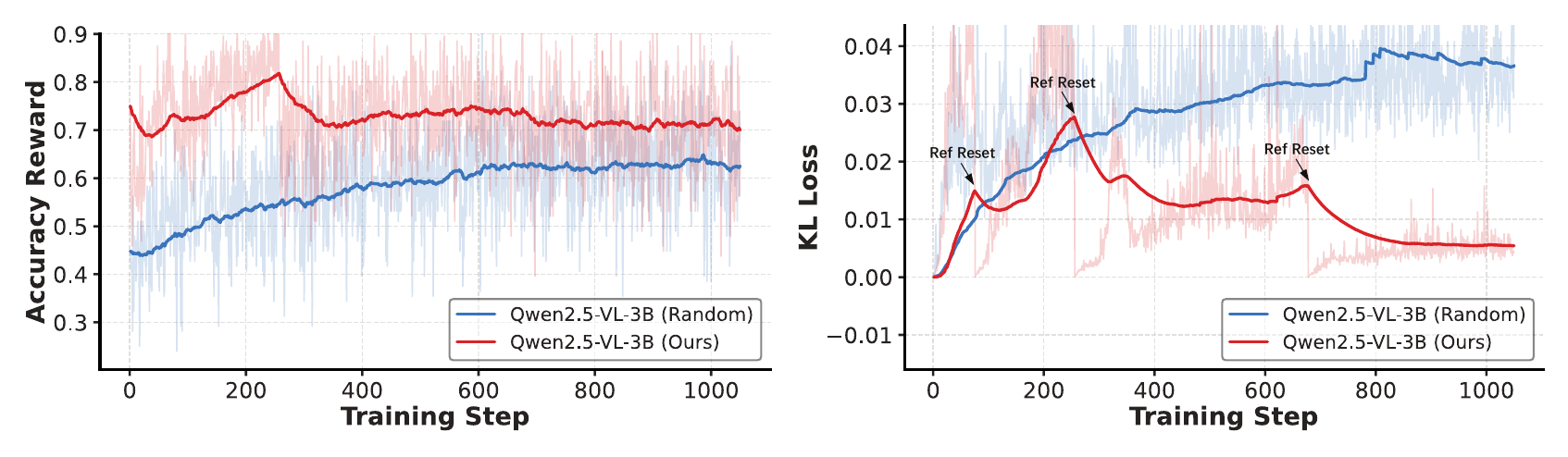}
    \caption{Training dynamics under \ours curriculum scheduling and randomly shuffled data. (Left) Accuracy reward. (Right) KL loss.}
    \label{fig:training_dynamics}
\end{figure}

\subsection{Training Comparison with Shuffled Data}
Figure~\ref{fig:training_dynamics} compares training dynamics between \ours and training with randomly shuffled data. We observe that the curriculum scheduling in \ours enables the model to achieve higher average accuracy rewards through better alignment between model capability and sample difficulty. Furthermore, the adaptive reference strategy reduces average KL loss, preventing over-alignment with the base model and improving reasoning capability. These benefits ultimately result in superior performance on the test set.

\begin{figure}[ht]
    \centering
    \includegraphics[width=1.0\linewidth]{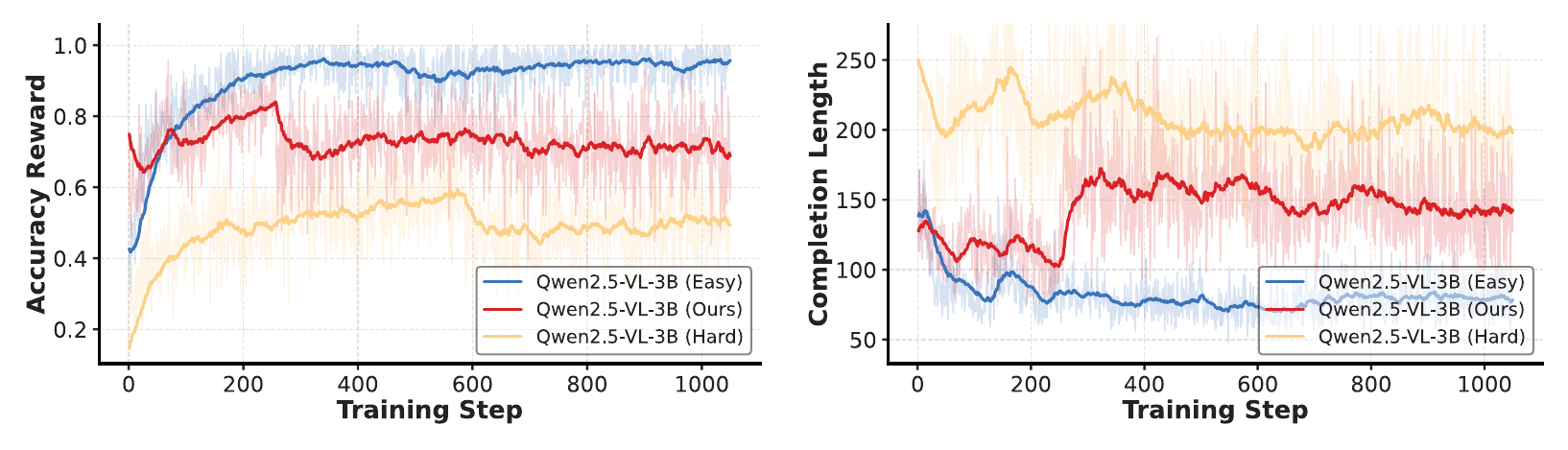}
    \caption{Reward and completion length during training with different difficulty distributions using Qwen2.5-VL-3B.}
    \vspace{-2mm}
    \label{fig:easy_hard}
\end{figure}

\subsection{Difficulty Distribution}
We further compare two alternative settings that exclusively use easy or hard samples. Specifically, we sample 10K training instances from \(\mathcal{G}_3\) and \(\mathcal{G}_1\), and train the model accordingly. We denote these as \ours(easy) and \ours(hard) in Table~\ref{tab:main_result}. Both variants underperform compared to the default difficulty setting, demonstrating the necessity of progressive difficulty data.

Figure~\ref{fig:easy_hard} provides further analysis. Training on easy data yields high rewards but fails to develop deeper reasoning capabilities, evidenced by shorter reasoning lengths that decrease during training. In contrast, training solely on hard data produces longer reasoning but often fails to reach correct answers, resulting in persistently low average rewards. With the default distribution, the model trains on appropriately challenging samples, maintaining high data utilization while steadily increasing reasoning length as harder data is gradually introduced.

\begin{table}[ht]
    \centering
    \small
    \setlength{\tabcolsep}{1.5mm}
        \begin{tabular}{l c c c c c c }
            \toprule
            \multirow{2}{*}{\textbf{Model}} & \textbf{Dyna} & \textbf{Math} & \textbf{Math-} & \textbf{Math} & \textbf{Logic} & \multirow{2}{*}{\textbf{Avg.}} \\
        & \textbf{Math} & \textbf{Vista} & \textbf{V} & \textbf{Verse} & \textbf{Vista} & \\
            \midrule
            \ours & \textbf{48.10} & \textbf{66.50} & \textbf{23.70} & \textbf{40.67} & \textbf{40.09} & \textbf{43.81} \\
            - SparseKL & 47.26 & 65.60 & 22.43 & 38.68 & 38.15 & 42.42 \\
            - Reset Ref & 44.65 & 63.90 & 22.46 & 37.95 & 38.93 & 41.58 \\
            - Revisiting & 46.26 & 65.60 & 22.63 & 36.18 & 38.03 & 41.74 \\
            - KL & 45.63 & 64.10 & 21.21 & 38.36 & 36.02 & 41.06 \\
            \bottomrule 
        \end{tabular}
    \caption{Ablation results on mathematical reasoning benchmarks using Qwen2.5-VL-3B.}
    
    \label{tab:ablation_math}
\end{table}
\begin{table}[ht]
    \centering
    \small
    \renewcommand{\arraystretch}{0.8}
    \begin{tabular}{l c c}
    \toprule
    \textbf{Model} & \textbf{\# Revisit} & \textbf{\# Degradation} \\
    \midrule
    Qwen2.5-VL-3B & 5528 & 1048 \\
    Qwen2.5-VL-7B & 5854 & 790 \\
    \bottomrule
    \end{tabular}
    \caption{Counts of revisits and reward degradations.}
    \vspace{-1.5mm}
    \label{tab:previous}
\end{table}
\begin{table}[ht]
    \centering
    \small
    \begin{tabular}{l|cc}
        \toprule
        \textbf{Base Model} & \textbf{Method} & \textbf{Avg.} \\
        \midrule
        \multirow{2}{*}{Qwen2.5-VL-3B} 
                                 & naive CL & 41.24 \\
                                 & AdaCuRL & \textbf{43.81} \\
        \midrule
        \multirow{2}{*}{Qwen2.5-VL-7B} 
                                 & naive CL  & 48.14 \\
                                 & AdaCuRL & \textbf{49.88} \\
        \midrule
        \multirow{2}{*}{Qwen2.5-Math-1.5B} 
                                 & naive CL & 30.22 \\
                                 & AdaCuRL & \textbf{32.32} \\
        \midrule
        \multirow{2}{*}{Qwen2.5-Math-7B} 
                                 & naive CL & 40.46 \\
                                 & AdaCuRL & \textbf{44.16} \\
        \bottomrule
    \end{tabular}
    \caption{Ablation results on different training scheduler. We provide the complete evaluation results in the Appendix~\ref{app_exp}.}
    \vspace{-2mm}
    \label{tab:online_comp}
\end{table}

\subsection{Ablation Study}

\noindent\textbf{Design of KL Divergence.} 
We evaluate two KL-related mechanisms, including SparseKL and Adaptive Ref.
Results in Tables~\ref{tab:ablation_math} and~\ref{tab:main} show that disabling either component degrades performance. We further observe that completely removing the KL divergence term from the loss results in a substantial performance drop. This is likely because revisiting earlier data amplifies overfitting to simpler samples, highlighting the necessity of the KL term in \ours. 

\noindent\textbf{Revisiting Historical Data.} 
\ours revisits historical samples by merging the next bucket and resetting the training data, which helps mitigate forgetting.
In this section, we analyze the forgetting issue and investigate an alternative strategy that keeps only the latest bucket without revisiting historical data. Table~\ref{tab:previous} shows statistics on the frequency of average group-reward decreases when previously seen samples were revisited, suggesting that training on harder samples can degrade performance on easier ones. Quantitative results in Tables~\ref{tab:main} and \ref{tab:ablation_math} show that appropriately revisiting past data further boosts performance.

\noindent\textbf{Dynamic Training Scheduler.}
\ours updates buckets dynamically based on average rewards during training. We also evaluate a naive curriculum strategy that processes samples from easy to hard using predefined buckets, without considering model feedback. As shown in Table~\ref{tab:online_comp}, this approach consistently underperforms \ours across all models, highlighting the limitations of fixed schedules that overlook the model’s evolving capabilities. 

More analyses are provided in the Appendix \ref{app_exp}.

\section{Related Work}
\subsection{Reasoning-oriented Reinforcement Learning} 
Reasoning for LLMs remains a central focus \cite{wang2024exploring, saparov2022language, xiong2025hs, wang2025position}. 
CoT Prompting~\cite{zhang2024improve, yao2023tree} guides models to reason step-by-step, while CoT Finetuning \cite{dong2025insight,xu2024llava} fine-tunes models on large-scale CoT datasets. DeepSeek-R1 \cite{guo2025deepseek} demonstrates that RL can spontaneously induce strong reasoning abilities, reducing the need for extensive CoT data. 
However, since MLLMs typically possess limited initial reasoning skills, applying RL directly yields minimal improvements.
This motivates studies \cite{yang2025r1, huang2025vision} to  distill CoT
data from DeepSeek-R1 or other reasoning-oriented models for SFT before RL, while \citet{huang2025boosting} provides expert reasoning chains during RL to solve hard problems. However, these methods overlook the alignment between model capability and sample difficulty.

\subsection{Curriculum Learning for RL} 
Curriculum learning (CL)~\cite{bengio2009curriculum} trains models from easy to hard and is now broadly used in RL~\cite{zhou2020uncertainty, wang2023efficienttrain}.
\citet{deng2025boosting} defines difficulty based on answer types, which fails to capture the model's intrinsic perception of difficulty. 
Other works~\cite{team2025kimi, deng2025boosting} employ fixed curricula without incorporating feedback from the model. \citet{shi2025efficient} estimate problem difficulty using expert models and propose an adaptive scheduler, however their method lacks historical data revisiting and does not address the degradation problem.
In contrast, \ours dynamically schedules samples based on model feedback and incorporates historical data revisiting to prevent performance degradation on early data. Finally, through a designed KL loss computation, the model avoids Policy Degradation when learning signals are absent.

\section{Conclusion}
This work tackles the challenges of Gradient Starvation and Policy Degradation in GRPO training caused by random data sampling. We propose \ours, a curriculum RL approach that dynamically adjusts training difficulty based on the model’s mastery of current samples. It also incorporates historical data replay and a meticulously designed KL divergence term to prevent reasoning deterioration. Without relying on external models or CoT datasets, \ours achieves significantly higher accuracy than random sampling on both multimodal and unimodal tasks using the same data. These results underscore the potential of curriculum learning in reasoning-oriented reinforcement learning.

\bibliography{aaai2026}

% appendix
% \input{sec/7_appendix}
% \clearpage
\appendix
\section{HyperParameters}
\label{app_hyp}
\subsection{Baseline}

\begin{table}[ht]
\centering
\resizebox{1.0\linewidth}{!}{
\begin{tabular}{l|rc}
\toprule
Model & Config & Value \\
\cmidrule(lr){1-3}

\multirow{9}{*}{Qwen2.5-VL}
&max prompt length & 1024 \\
&max completion length & 1024 \\
&temperature & 1.0 \\
&learning rate & 1e-6 \\
&learning schedule & Linear \\
&global batchsize & 8 \\
&gradient accumulation & 1 \\
&num generations & 6 \\
&epoch & 1 \\

\midrule

\multirow{11}{*}{Qwen2.5-Math}
&max prompt length & 512 \\
&max completion length & 3584 \\
&temperature & 0.7 \\
&learning rate & 1e-6 \\
&lr schedule & cosine\_with\_min\_lr \\
&min lr rate & 0.1 \\
&warmup ratio & 0.1 \\
&global batchsize & 96 (1.5B), 48 (7B) \\
&gradient accumulation & 1 (1.5B), 2 (7B) \\
&num generations & 6 \\
&epoch & 1 \\

\bottomrule
\end{tabular}
}
\caption{Training hyperparameters for MLLM and LLM GRPO baselines.}
\label{tab:hyp_baseline} 
\end{table}

The training hyperparameters for the GRPO baseline models are listed in Table~\ref{tab:hyp_baseline}. We follow the basic settings from R1-V for the hyperparameters in the table.

\subsection{\ours}
\begin{table}[ht]
\centering
\resizebox{1.0\linewidth}{!}{
\begin{tabular}{l|rc}
\toprule
Model & Config & Value \\ 
\midrule

\multirow{12}{*}{Qwen2.5-VL} 
& max prompt length & 1024 \\ 
& max completion length & 1024 \\
&temperature & 1.0 \\
& learning rate & 1e-6 \\ 
& learning schedule & Linear \\ 
& global batchsize & 16 \\ 
& gradient accumulation & 1 \\ 
& num generations & 6 \\ 
& max steps & 1050 \\ 
& \(T_f\) & 64 \\ 
& \(M\) & 512 \\ 
& buckets number \(K\) & 4 \\ 

\midrule

\multirow{14}{*}{Qwen2.5-Math} 
& max prompt length & 512 \\ 
& max completion length & 3584 \\
&temperature & 0.7 \\
& learning rate & 1e-6 \\ 
& lr schedule & cosine\_with\_min\_lr \\ 
& min lr rate & 0.1 \\ 
& warmup ratio & 0.1 \\ 
& global batchsize & 96 (1.5B), 48 (7B) \\ 
& gradient accumulation & 1 (1.5B), 2 (7B) \\ 
& num generations & 6 \\ 
& max steps & 650 \\ 
& \(T_f\) & 64 \\ 
& \(M\) & 512 \\ 
& buckets number \(K\) & 3 \\ 

\bottomrule
\end{tabular}
}
\caption{Training hyperparameters for MLLM and LLM using the \ours.}
\label{tab:hyp_ours} 
\end{table}

The training hyperparameters for training MLLMs and LLMs using \ours are listed in Table~\ref{tab:hyp_ours}. The format reward cutoff step $T_f$ is set to 64, as we observe that the format reward typically converges around 32 steps during GRPO baseline training. We set the competence score update step $M$ to 512 to balance training efficiency and accurately reflect the model's mastery of the current training difficulty.

We study the impact of the number of curriculum buckets \(K\). A larger \(K\) leads to more frequent training on previously seen samples, potentially causing overfitting to easier samples due to excessive repetition. In contrast, a smaller \(K\) increases the risk of forgetting. Additionally, since the reference model is reset between buckets, a smaller \(K\) results in stronger alignment with the base model due to fewer resets. As shown in Table~\ref{tab:buckets_math}, \ours demonstrates robustness to moderate bucket counts (e.g., 3, 4, and 5 buckets), while extreme values (e.g., 1 or 10 buckets) lead to performance degradation. Setting \(K = 5\) achieves the best trade-off among these factors, yielding the highest overall performance.
The number of buckets is set to 3 for LLMs due to the relatively smaller amount of training data, and we do not perform additional search for the optimal bucket count.

\begin{table}[ht]
    \centering
    \small
    \setlength{\tabcolsep}{1mm}
    
    \resizebox{1.0\linewidth}{!}{
    \begin{tabular}{l c c c c c c }
        \toprule
        \textbf{Buckets Num} & \textbf{DynaMath} & \textbf{MathVista} & \textbf{Math-V} & \textbf{MathVerse} & \textbf{LogicVista} & \textbf{Avg.} \\
        \midrule
        1 buckets & 43.54 & 65.10 & 23.15 & 38.06 & 38.94 & 41.76  \\
        3 buckets & 48.46 & \textbf{66.70} & 22.96 & 37.96 & 40.04 & 43.22  \\
        4 buckets & 48.10 & 66.50 & \textbf{23.70} & \textbf{40.67} & 40.09 & \textbf{43.81} \\
        5 buckets & \textbf{49.04} & 65.7 & 22.46 & 39.30 & \textbf{40.94} &  43.49 \\
        10 buckets & 47.10 & 65.00 & 22.20 & 38.83 & 37.58 & 42.14 \\

        \bottomrule 
    \end{tabular}
    }
        \caption{Results with different numbers of buckets on mathematical reasoning benchmarks (Qwen2.5-VL-3B).} 
    \label{tab:buckets_math}
\end{table}

\begin{table}[ht]
\centering
\begin{tabular}{l|rc}
\toprule
Model & Config & Value \\
\midrule

\multirow{4}{*}{Qwen2.5-VL-3B}
&max token & 2048 \\
&temperature & 1.6 \\
&top\_p & 0.95 \\
&top\_k & 50 \\

\midrule

\multirow{4}{*}{Qwen2.5-VL-7B}
&max token & 4096 \\
&temperature & 1.6 \\
&top\_p & 0.95 \\
&top\_k & 50 \\

\midrule

\multirow{4}{*}{Qwen2.5-Math-1.5B}
&max token & 2048 \\
&temperature & 0.7 \\
&top\_p & 0.8 \\
&top\_k & 50 \\

\midrule

\multirow{4}{*}{Qwen2.5-Math-7B}
&max token & 4096 \\
&temperature & 0.2 \\
&top\_p & 0.8 \\
&top\_k & 50 \\

\bottomrule
\end{tabular}
\caption{Training hyperparameters for MLLM and LLM GRPO baselines.}
\label{tab:hyp_diff} 
\end{table}

\subsection{Difficulty Estimation}
Table~\ref{tab:hyp_diff} presents the generation parameters used for coarse-grained and fine-grained difficulty estimation in \ours, where we manually adjust these parameters for different models to control output diversity and avoid identical or repetitive generations.

\section{Details of Training Data}
Table~\ref{tab:training data} presents the dataset partitioning results obtained after coarse-grained difficulty estimation for the Qwen2.5-VL-3B-Instruct and Qwen2.5-VL-7B-Instruct models, along with the final datasets constructed through balanced sampling at the dataset level. In the table, Re-\ours indicates the results of coarse-grained difficulty estimation performed by the policy model after one round of curriculum reinforcement learning.

Furthermore, the results in the table also demonstrate the necessity of coarse-grained difficulty estimation. For example, for the Qwen2.5-VL-3B model, the sample counts in $\mathcal{G}1$, $\mathcal{G}_2$, and $\mathcal{G}_3$ are almost uniformly distributed across the entire dataset, meaning that if we directly randomly sample 10K samples for fine-grained partitioning, the resulting difficulty distribution would also be uniform, containing a large number of simple samples that cannot effectively enhance the model's reasoning capability.

\begin{figure}[ht]
    \centering
    \includegraphics[width=0.8\linewidth]{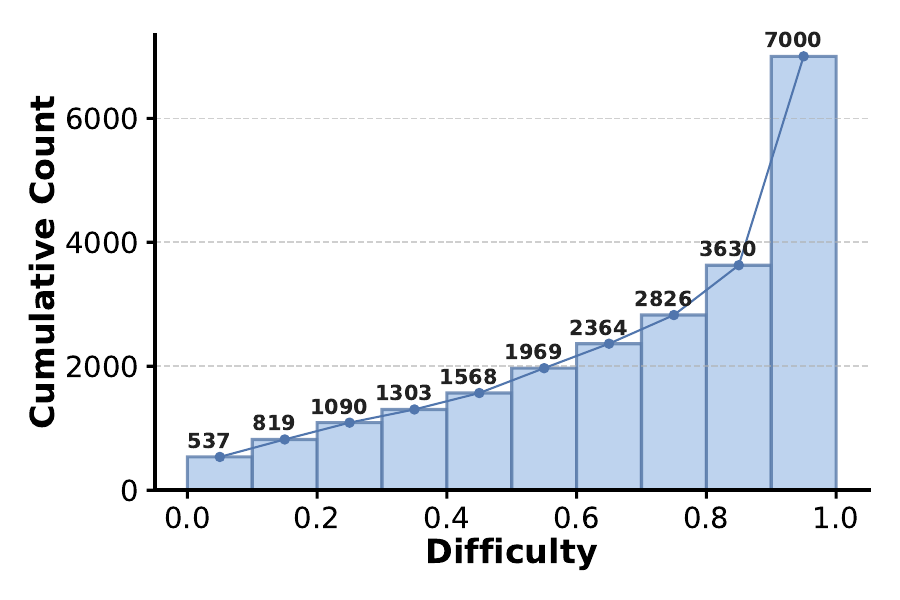}
    \caption{
     Difficulty distribution of the sampled dataset after fine estimation with the Qwen2.5-Math-1.5B model.
}
    \label{fig:distribution_llm}
\end{figure}

For large language models (LLMs), the Open-RS dataset we utilized contains only 7K samples, thus eliminating the need for a coarse-to-fine difficulty estimation strategy. Instead, we directly perform fine-grained difficulty estimation. Figure~\ref{fig:distribution_llm} illustrates the difficulty estimation results for the Qwen2.5-Math-1.5B model. It can be observed that the difficulty distribution of this dataset skews towards the harder side, which is consistent with our findings in multimodal experiments: a higher proportion of challenging data can enhance the reasoning capabilities of the model. Notably, the Open-RS dataset itself provides difficulty annotations, including 1K \emph{easy} samples and \emph{6K} hard samples. These annotations align well with the results shown in the figure, but our method offers a more fine-grained difficulty estimation, which facilitates subsequent reinforcement curriculum learning.

\begin{table*}[ht]
    \centering
    \resizebox{\textwidth}{!}{
    \begin{tabular}{l l c c c c c c c c c c c c c c}
        \toprule
        \textbf{Model} & \textbf{Group} &
        \textbf{clever} & \textbf{clever\_math} & \textbf{geo3k} & \textbf{geomverse} &
        \textbf{geoqa\_plus} & \textbf{iconqa} & \textbf{super\_clever} & \textbf{tabmwp} &
        \textbf{unigeo} & \textbf{wemath} & \textbf{SceMQA} & \textbf{polymath} & \textbf{GEOS} & \textbf{Sum} \\
        \midrule
        % ----------------------------------------------------------
        \multirow{6}{*}{Qwen2.5-VL-3B}
            & $\mathcal{G}_1$        & 619 & 1142 & 1324 & 558 & 25678 & 3318 & 1135 & 786 & 474 & 1105 & 41 & 3070 & 27 & 39277 \\
            & $\mathcal{G}_2$        & 566 & 246 & 926 & 554 & 19502 & 6824 & 640 & 804 & 473 & 545 & 19 & 1379 & 14 & 32492 \\
            & $\mathcal{G}_3$        & 12277 & 1560 & 151 & 574 & 3880 & 10423 & 1823 & 888 & 165 & 90 & 3 & 149 & 5 & 31988 \\
            \cmidrule(lr){2-16}
            & Samples $\mathcal{G}_1$ & 448 & 448 & 448 & 448 & 448 & 448 & 448 & 448 & 448 & 448 & 41 & 448 & 27 & 4996 \\
            & Samples $\mathcal{G}_2$ & 272 & 246 & 272 & 272 & 272 & 272 & 272 & 272 & 272 & 272 & 19 & 272 & 14 & 2999 \\
            & Samples $\mathcal{G}_3$ & 205 & 205 & 151 & 205 & 205 & 205 & 205 & 205 & 165 & 90 & 3 & 149 & 5 & 1998 \\
        \midrule
        % ----------------------------------------------------------
        \multirow{6}{*}{Qwen2.5-VL-3B (Re-\ours)}
            & $\mathcal{G}_1$        & 74 & 849 & 1017 & 451 & 19221 & 2203 & 1150 & 651 & 262 & 960 & 30 & 2469 & 21 & 29358 \\
            & $\mathcal{G}_2$        & 150 & 135 & 985 & 408 & 19871 & 4050 & 437 & 627 & 434 & 535 & 26 & 1596 & 16 & 29270 \\
            & $\mathcal{G}_3$        & 13238 & 1964 & 399 & 827 & 9968 & 14312 & 2011 & 1200 & 416 & 245 & 7 & 353 & 9 & 44949 \\
            \cmidrule(lr){2-16}
            & Samples $\mathcal{G}_1$ & 74 & 520 & 520 & 451 & 520 & 520 & 520 & 520 & 262 & 520 & 30 & 520 & 21 & 4998 \\
            & Samples $\mathcal{G}_2$ & 150 & 135 & 297 & 297 & 297 & 297 & 297 & 297 & 297 & 297 & 26 & 297 & 16 & 3000 \\
            & Samples $\mathcal{G}_3$ & 194 & 194 & 52 & 194 & 194 & 194 & 194 & 194 & 194 & 194 & 8 & 194 & 0 & 2000 \\
        \midrule
        % ----------------------------------------------------------
        \multirow{6}{*}{Qwen2.5-VL-7B}
            & $\mathcal{G}_1$        & 210 & 912 & 1774 & 185 & 21424 & 1917 & 1038 & 529 & 343 & 1001 & 46 & 3350 & 30 & 32759 \\
            & $\mathcal{G}_2$        & 921 & 248 & 575 & 189 & 20452 & 3982 & 532 & 411 & 525 & 538 & 9 & 1037 & 16 & 29435 \\
            & $\mathcal{G}_3$        & 12331 & 1788 & 52 & 1312 & 7184 & 14666 & 2028 & 1538 & 244 & 201 & 8 & 211 & 0 & 41563 \\
            \cmidrule(lr){2-16}
            & Samples $\mathcal{G}_1$ & 210 & 523 & 523 & 185 & 523 & 523 & 523 & 523 & 343 & 523 & 46 & 523 & 30 & 4998 \\
            & Samples $\mathcal{G}_2$ & 282 & 248 & 282 & 189 & 282 & 282 & 282 & 282 & 282 & 282 & 9 & 282 & 16 & 3000 \\
            & Samples $\mathcal{G}_3$ & 194 & 194 & 52 & 194 & 194 & 194 & 194 & 194 & 194 & 194 & 8 & 194 & 0 & 2000 \\
        \midrule
        % ----------------------------------------------------------
        \multirow{6}{*}{Qwen2.5-VL-7B (Re-\ours)}
            & $\mathcal{G}_1$        & 38 & 832 & 963 & 144 & 9945 & 1197 & 990 & 455 & 134 & 648 & 33 & 2631 & 25 & 18035 \\
            & $\mathcal{G}_2$        & 44 & 99 & 975 & 113 & 13809 & 1938 & 385 & 310 & 379 & 471 & 18 & 1340 & 16 & 19897 \\
            & $\mathcal{G}_3$        & 13380 & 2017 & 463 & 1429 & 25306 & 17430 & 2223 & 1713 & 599 & 621 & 12 & 627 & 5 & 65825 \\
            \cmidrule(lr){2-16}
            & Samples $\mathcal{G}_1$ & 38 & 596 & 596 & 144 & 596 & 596 & 596 & 455 & 134 & 596 & 33 & 596 & 25 & 5001 \\
            & Samples $\mathcal{G}_2$ & 44 & 99 & 400 & 113 & 400 & 400 & 400 & 310 & 400 & 400 & 18 & 400 & 16 & 3000 \\
            & Samples $\mathcal{G}_3$ & 180 & 180 & 180 & 180 & 180 & 180 & 180 & 180 & 180 & 180 & 12 & 180 & 5 & 1997 \\
        \bottomrule
    \end{tabular}
    }
    \caption{The coarse-grained difficulty estimation results of different models and the training data sampling results for each dataset}
    \label{tab:training data}
\end{table*}

\begin{algorithm}[ht]
\caption{Curriculum-Based Reinforcement Learning}
\label{alg:algorithm}
\begin{algorithmic}[1]
\STATE \textbf{Input:} Initial policy model $\pi_\theta$, sorted dataset $\mathcal{D}$, number of buckets $K$, format reward cutoff step $T_f$, training step $T \gets 0$, decay factor $\gamma$, competence score $cs \gets 0$, reward accumulation length $M$
\STATE \textbf{Init:} Curriculum set $\mathcal{D}_c \gets \emptyset$, reward buffer $\mathcal{R}_b \gets \emptyset$
\STATE Partition $\mathcal{D}$ into equal buckets $\{\mathcal{B}_1, \mathcal{B}_2, ..., \mathcal{B}_K\}$
\STATE $\mathcal{D}_c \gets \mathcal{B}_1$
\WHILE{$\mathcal{B}_K \not\subset \mathcal{D}_c$ and samples in $\mathcal{B}_K$ not fully trained}
    \STATE Sample batch $X \sim \mathcal{D}_c$
    \STATE Generate $G \gets \pi_\theta(X)$
    \IF{$T < T_f$}
        \STATE $R \gets \text{Acc}(X, G) + \text{Format}(X, G)$
    \ELSE
        \STATE $R \gets \text{Acc}(X, G)$
    \ENDIF
    \STATE Update policy $\pi_\theta$ using Eq.~\eqref{eq:grpo-loss}
    \STATE Append $R$ to $\mathcal{R}_s$
    \IF{$|\mathcal{R}_s| \ge M$ and $T \ge T_f$}
        \STATE Compute $\bar{r} \gets$ average of $\mathcal{R}_s$
        \STATE Update competence score:
        \[
        cs \gets cs + (\bar{r} - 0.5) \cdot \max(1 - cs, \gamma)
        \]
        \IF{$g \ge \frac{k - 1}{K}$}
            \STATE $\mathcal{D}_c \gets \mathcal{D}_c \cup \mathcal{B}_k$, reset $\mathcal{R}_s$, set $cs \gets \frac{k - 1}{K}$ \label{alg:line19}
        \ENDIF
    \ENDIF
    \STATE Check whether $R$ is all 0 or 1 (i.e., whether $\hat{A}_i$ is a zero vector), and compute the GRPO loss as:
    \[
        \mathcal{L}_{\text{GRPO}}(\theta)
        = -\,\mathbb{E}_{i}\!\bigl[\rho_i\,\hat{A}_i\bigr]
        + \mathbb{I}\!\bigl[\hat{A}_i \neq 0\bigr]\,
          \beta\,
          \mathbb{E}_{i}\!\Bigl[
            \mathrm{KL}\!\bigl(\pi_\theta \,\|\, \pi_{\text{ref}}\bigr)
          \Bigr]
    \]
    \STATE $T \gets T + 1$
\ENDWHILE
\end{algorithmic}
\end{algorithm}

\section{Algorithm}
\label{app_algo}
The overall procedure of \ours is illustrated in Algorithm~\ref{alg:algorithm}.

\section{More Experimental Results}
\label{app_exp}
\begin{table}[ht]
    \centering
    \small
    \setlength{\tabcolsep}{1mm}    
    \resizebox{1.0\linewidth}{!}{
    \begin{tabular}{l c c c c c c }
        \toprule
        \textbf{Model} & \textbf{MMStar} & \textbf{MMMU} & \textbf{Hallu.} & \textbf{AI2D} & \textbf{MMVET} & \textbf{Avg.} \\
        \midrule
        \ours & \textbf{59.95} & 52.66 & \textbf{49.03} & \textbf{81.34} & \textbf{62.76} & \textbf{61.15} \\
        - SparseKL & 58.66 & 52.44 & 48.70 & 80.56 & 61.83 & 60.44 \\
        - Reset Ref & 56.87 & 51.00 & 48.69 & 80.86 & 62.20 & 59.92 \\
        - Revisiting & 57.20 & \textbf{52.88} & 46.86 & 78.85 & 59.38 & 59.03 \\
        - KL & 58.40 & 51.22 & 46.94 & 81.02 & 61.00 & 59.72 \\

        \bottomrule 
    \end{tabular}
    }
        \caption{Ablation results on general reasoning benchmarks (Qwen2.5-VL-3B)}  
    \label{tab:ablation_general}
\end{table}

\subsection{More Ablation Results}
We present ablation results on general reasoning benchmarks in Table~\ref{tab:ablation_general}.

\subsection{Design of KL divergence}
\begin{figure}[ht]
    \centering
    \includegraphics[width=0.9\linewidth]{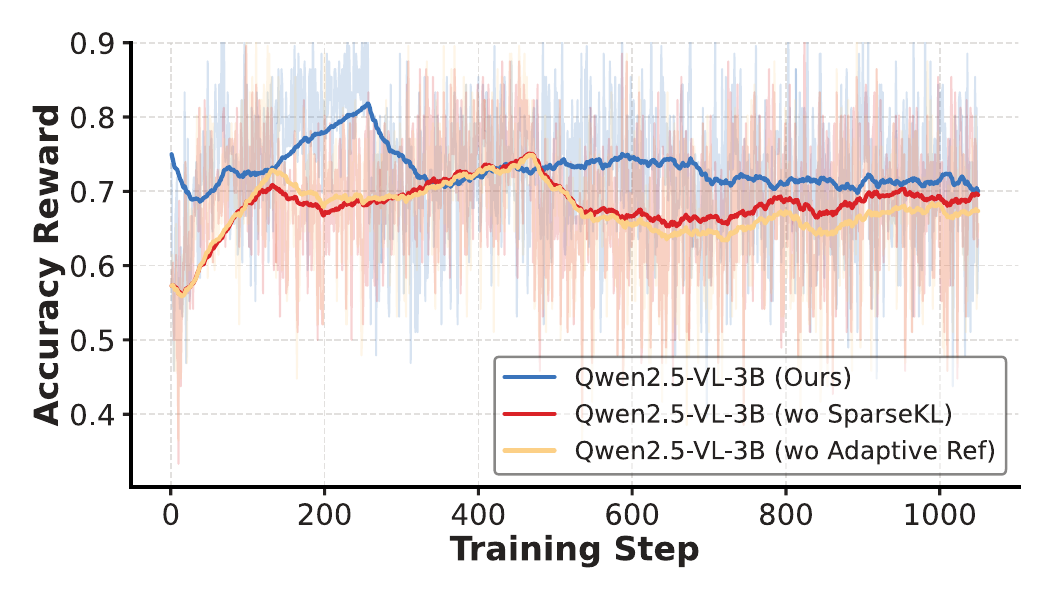}
    \caption{
    Reward variation with different KL divergence designs in Qwen2.5-VL-3B training.
}
    \label{fig:kl}
\end{figure}

We further analyze the impact of the KL divergence design in \ours on the RL process. As shown in Figure~\ref{fig:kl}, \ours achieves a higher average accuracy reward during training compared to its two variants that remove either Sparse KL or Adaptive Ref. This indicates that the KL divergence design in \ours effectively prevents inference degradation caused by excessive alignment of the policy model with the base model when learning signals are absent.

\begin{figure}[ht]
    \centering
    \includegraphics[width=0.95\linewidth]{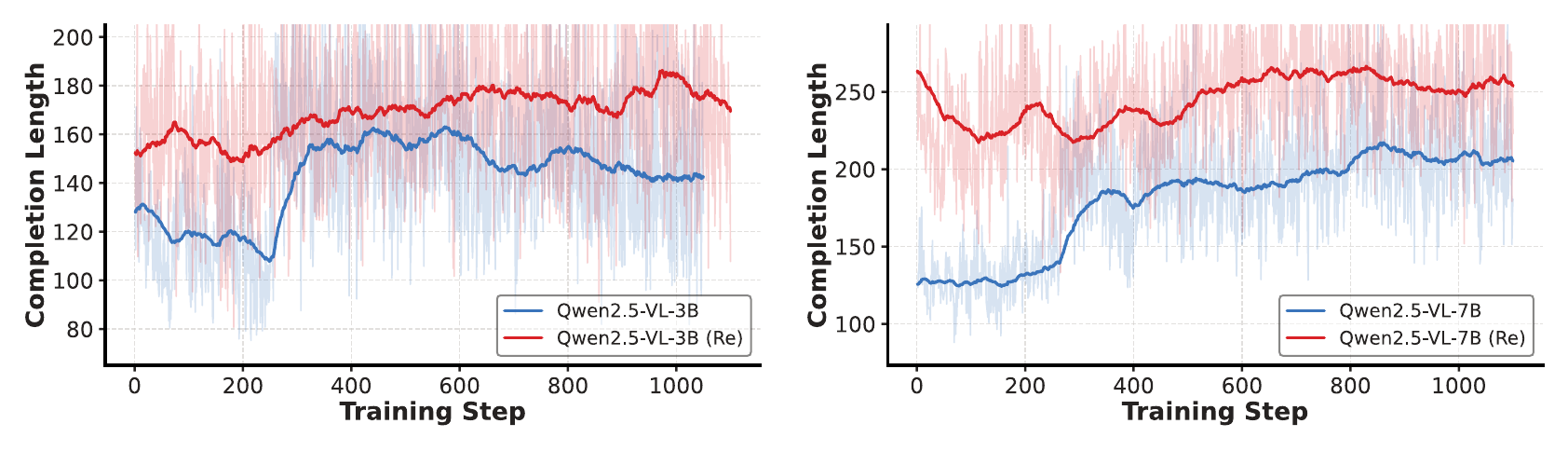}
    \caption{Comparison of completion length between \ours and Re-\ours.}
    \label{fig:recur}
\end{figure}

\subsection{Re-\ours}
Figure~\ref{fig:recur} demonstrates that during iterative training, the reasoning length of the policy model consistently exceeds that of the initial training round. This observation indicates that the updated policy model, through re-estimating difficulty, continues to identify and train on appropriately challenging samples from the large-scale dataset, thereby further enhancing its reasoning capability and data utilization efficiency.

\begin{table*}[ht]
    \centering
    \small
    \begin{tabular}{lccccc c}
        \toprule
        \textbf{Method} & \textbf{DynaMath} & \textbf{MathVista} & \textbf{MathVision} & \textbf{MathVerse} & \textbf{LogicVista} & \textbf{Avg.} \\
        \midrule
        \multicolumn{7}{c}{\textbf{\textit{Qwen2.5-VL-3B}}} \\
        \midrule
        naive CL & 42.35 & 64.6 & 23.21 & 37.15 & 38.91 & 41.24 \\
        \ours    & \textbf{48.10} & \textbf{66.50} & \textbf{23.70} & \textbf{40.67} & \textbf{40.09} & \textbf{43.81} \\
        \midrule
        \multicolumn{7}{c}{\textbf{\textit{Qwen2.5-VL-7B}}} \\
        \midrule
        naive CL & 52.00 & 68.40 & 26.77 & 47.94 & 45.63 & 48.14 \\
        \ours    & \textbf{55.10} & \textbf{70.40} & \textbf{27.07} & \textbf{48.75} & \textbf{48.10} & \textbf{49.88} \\
        \bottomrule
    \end{tabular}
    \caption{Ablation results of different training schedulers on MLLMs.}
    \label{tab:app_online_comp1}
\end{table*}

\begin{table*}[ht]
    \centering
    \small
    \begin{tabular}{lccccc c}
        \toprule
        \textbf{Method} & \textbf{AIME24} & \textbf{AMC23} & \textbf{MATH500} & \textbf{Minerva} & \textbf{Olymp.} & \textbf{Avg.} \\
        \midrule
        \multicolumn{7}{c}{\textbf{\textit{Qwen2.5-Math-1.5B}}} \\
        \midrule
        naive CL & 9.37 & 42.81 & 57.20 & 13.35 & 28.39 & 30.22 \\
        \ours    & \textbf{9.58} & \textbf{45.63} & \textbf{62.46} & \textbf{14.58} & \textbf{29.33} & \textbf{32.32} \\
        \midrule
        \multicolumn{7}{c}{\textbf{\textit{Qwen2.5-Math-7B}}} \\
        \midrule
        naive CL & 19.79 & 56.40 & 69.33 & 21.81 & 34.96 & 40.46 \\
        \ours    & \textbf{22.22} & \textbf{59.22} & \textbf{74.53} & \textbf{27.33} & \textbf{37.48} & \textbf{44.16} \\
        \bottomrule
    \end{tabular}
    \caption{Ablation results of different training schedulers on LLMs.}
    \label{tab:app_online_comp2}
\end{table*}

\subsection{Results of Different Training Schedulers}
We provide detailed comparisons between \ours and the naive curriculum scheduling algorithm across various benchmarks in Table~\ref{tab:app_online_comp1} and \ref{tab:app_online_comp2}. It can be observed that \ours outperforms the naive scheduling algorithm across all benchmarks, highlighting the importance of considering model feedback in curriculum updates.

\subsection{Statistical Significance Analysis}

We employ the Wilcoxon signed-rank test to evaluate the statistical significance of improvements achieved by \ours across four models. For multimodal models, we use the mathematical reasoning benchmark and general reasoning benchmark from the main text for statistical analysis, while for language models, we use the mathematical reasoning benchmark for statistics. We set the significance level $\alpha = 0.05$ as the threshold for statistical significance. Results are shown in Table~\ref{tab:significance_analysis}, where it can be observed that the p-values for all models are below this threshold, indicating that \ours significantly outperforms the baseline GRPO method with strong statistical evidence.

\begin{table}[ht]
\centering
\caption{Wilcoxon Signed-Rank Test Results}
\label{tab:significance_analysis}
\begin{tabular}{lcc}
\toprule
Model & Sample Size & p-value \\
\midrule
Qwen2.5-VL-3B & 10 & 0.001 \\
Qwen2.5-VL-7B & 10 & 0.005 \\
Qwen2.5-Math-1.5B & 5 & 0.031 \\
Qwen2.5-Math-7B & 5 & 0.031 \\
\bottomrule
\end{tabular}
\end{table}

\subsection{Statistical Significance Analysis}

We employ the Wilcoxon signed-rank test to evaluate the statistical significance of improvements achieved by \ours across four models. For multimodal models, we use the mathematical reasoning benchmark and general reasoning benchmark from the main text for statistical analysis, while for language models, we use the mathematical reasoning benchmark for statistics. We set the significance level $\alpha = 0.05$ as the threshold for statistical significance. Results are shown in Table~\ref{tab:significance_analysis}, where it can be observed that the p-values for all models are below this threshold, indicating that \ours significantly outperforms the baseline GRPO method with strong statistical evidence.

\section{Details of the Evaluation Benchmarks}
\subsection{Multimodal Reasoning Benchmark}
For MLLMs, we build two complementary benchmarks: one targeting mathematical reasoning and the other assessing general multimodal reasoning.

The \emph{mathematical reasoning benchmark} includes the following datasets:
\begin{itemize}[leftmargin=*]
    \item \textbf{DynaMath} — A large-scale benchmark with 5,010 questions designed to evaluate the robustness of multimodal mathematical reasoning in dynamic visual and textual contexts.
    \item \textbf{MathVista\_MINI} — A curated set of 1,000 test samples covering diverse mathematical and visual reasoning challenges.
    \item \textbf{MathVision} — A collection of 3,040 high-quality math problems drawn from real-world math competitions.
    \item \textbf{MathVerse\_MINI} — Designed to assess whether and to what extent MLLMs can truly understand visual diagrams in mathematical contexts.
    \item \textbf{LogicVista} — Targets the evaluation of integrated logical reasoning within visual environments.
\end{itemize}

The \emph{general reasoning benchmark} consists of the following:
\begin{itemize}[leftmargin=*]
    \item \textbf{MMStar} — An elite vision-dependent benchmark comprising 1,500 challenge samples requiring fine-grained visual reasoning.
    \item \textbf{MMMU} — Covers a broad range of multi-discipline tasks that demand college-level subject knowledge and deliberate multimodal reasoning.
    \item \textbf{HallusionBench} — A diagnostic benchmark of 1,149 questions designed to evaluate models’ ability to reason over image-grounded contexts.
    \item \textbf{AI2D} — Contains over 5,000 grade-school science diagrams with more than 150,000 richly annotated elements.
    \item \textbf{MMVET} — A comprehensive benchmark assessing six core multimodal capabilities: OCR, visual grounding, commonsense reasoning, visual recognition, inference, and spatial understanding.
\end{itemize}

\subsection{Language Modality Reasoning Benchmark}
For LLMs, we adopt standard mathematical reasoning benchmarks.
\begin{itemize}[leftmargin=*]
    \item \textbf{AIME24} — This dataset contains problems from the 2024 American Invitational Mathematics Examination (AIME), a prestigious high school mathematics competition known for its challenging math problems. The dataset includes a total of 30 records.
    \item \textbf{AMC23} — Including problems from the 2023 American Mathematics Competitions.
    \item \textbf{Math500} — A challenging high school math competition dataset consisting of 500 problems across seven subjects.
    \item \textbf{Minerva} — A quantitative reasoning benchmark containing approximately 500 challenging mathematical problems that require multi-step reasoning and real-world applications.
    \item \textbf{Olympiadbench} — An Olympiad-level bilingual multimodal scientific benchmark comprising 8,476 math and physics problems.
\end{itemize}

\end{document}